\documentclass[letterpaper, 10 pt, conference]{ieeeconf}
\usepackage{amsmath,amsfonts}
\usepackage{array}
\usepackage{textcomp}
\usepackage{stfloats}
\usepackage{url}
\usepackage{verbatim}
\usepackage{graphicx}
\usepackage{cite}
\usepackage{xcolor}
\usepackage{subcaption}
\usepackage{mathtools}
\usepackage{amssymb}
\usepackage{tabulary}
\usepackage{booktabs}
\usepackage[ruled,linesnumbered]{algorithm2e}
\usepackage{setspace}
\usepackage{multirow}
\usepackage{enumerate}
\makeatletter
\let\NAT@parse\undefined
\makeatother
\usepackage{hyperref}
\usepackage{makecell}

\IEEEoverridecommandlockouts
\title{\LARGE \bf
You Only Estimate Once: Unified, One-stage,
Real-Time Category-level Articulated Object 6D Pose
Estimation for Robotic Grasping
}

\author{Jingshun Huang$^{1*}$\quad Haitao Lin$^{2*}$\quad  Tianyu Wang$^{1}$\quad Yanwei Fu$^{1}$ \quad  Yu-Gang Jiang$^{1}$\quad  Xiangyang Xue$^{1}$ \\
\thanks{$*$ indicates equal contribution.}
\thanks{$^{1}$Fudan University. jshuang23@m.fudan.edu.cn.}
\thanks{$^{2}$Tencent Robotics X Lab, Shenzhen, China.}
}

\hypersetup{
colorlinks=true,
linkcolor=black,
citecolor=black
}

\makeatletter
\let\@oldmaketitle\@maketitle

\begin{document}

\maketitle

\begin{abstract}

    This paper addresses the problem of category-level pose estimation for articulated objects in robotic manipulation tasks. Recent works have shown promising results in estimating part pose and size at the category level. However, these approaches primarily follow a complex multi-stage pipeline that first segments part instances in the point cloud and then estimates the Normalized Part Coordinate Space (NPCS) representation for 6D poses. These approaches suffer from high computational costs and low performance in real-time robotic tasks. 
    To address these limitations, we propose YOEO, a single-stage method that simultaneously outputs instance segmentation and NPCS representations in an end-to-end manner. 
    We use a unified network to generate point-wise semantic labels and centroid offsets, allowing points from the same part instance to vote for the same centroid. We further utilize a clustering algorithm to distinguish points based on their estimated centroid distances. Finally, we first separate the NPCS region of each instance. Then, we align the separated regions with the real point cloud to recover the final pose and size. 
    Experimental results on the GAPart dataset demonstrate the pose estimation capabilities of our proposed single-shot method. We also deploy our synthetically-trained model in a real-world setting, providing real-time visual feedback at 200Hz, enabling a physical Kinova robot to interact with unseen articulated objects. This showcases the utility and effectiveness of our proposed method \footnote[2]{Project webpage. \url{https://shanehuanghz.github.io/YOEO/}}.

\end{abstract}

\section{INTRODUCTION}

\begin{figure}
    \centering
    \includegraphics[width=0.5\textwidth]{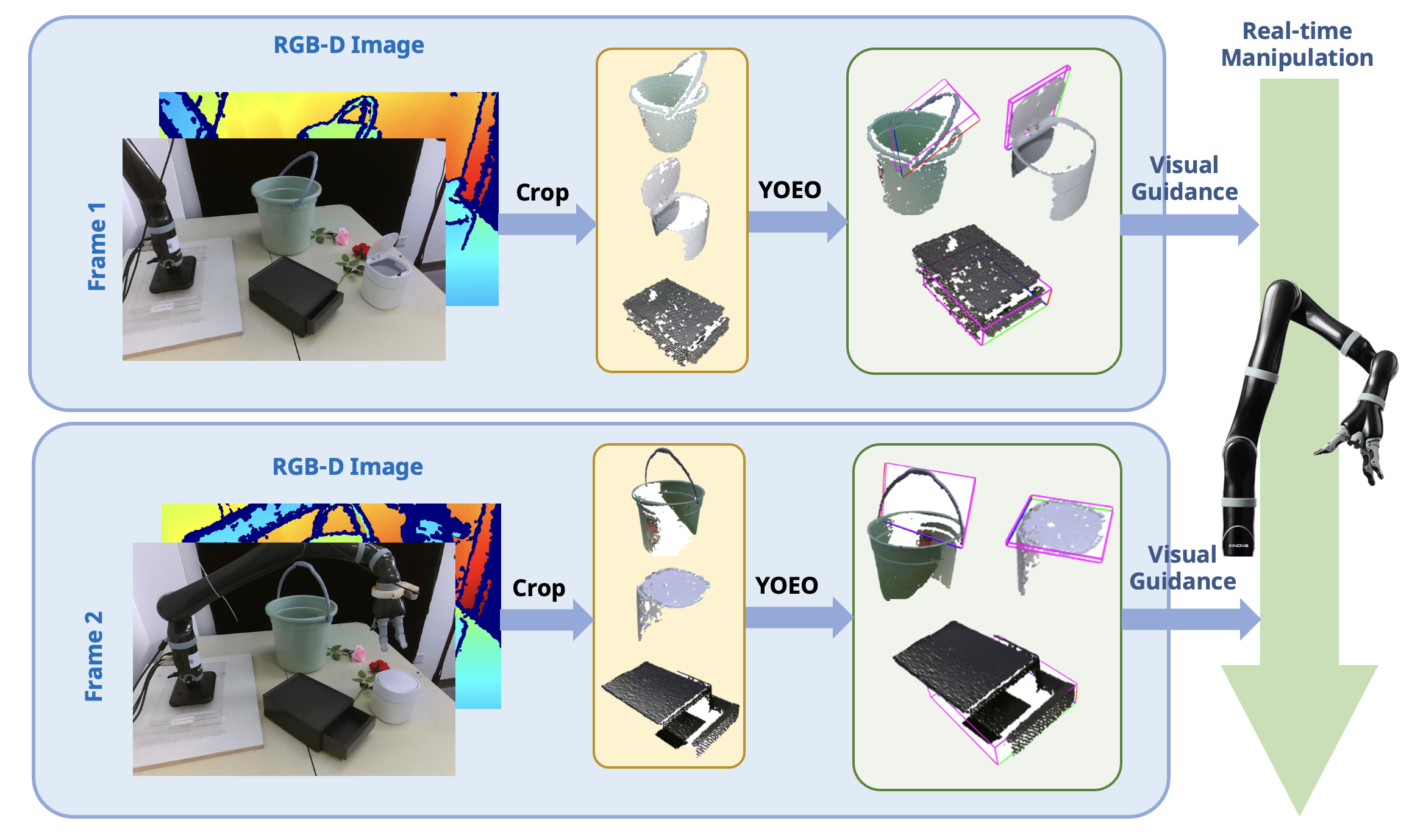}
       \vspace{-0.1in}
    \caption{\textbf{Overview.} We propose a unified, single-stage method for articulated object 6D pose estimation named YOEO, which enables real-time robotic manipulation.     \label{fig:teaser} }
       \vspace{-0.2in}
\end{figure}

Accurately estimating the state information of objects is crucial for robots before undertaking motion planning in various grasping and manipulation tasks~\cite{wang2019densefusion, lin2022know, lin2022sar, wen2023foundationpose}, as shown in Fig~\ref{fig:teaser}. Recent research~\cite{chang2015shapenet} has made significant progress in estimating the state of rigid bodies from single images. However, estimating the state information of non-rigid bodies remains challenging due to their complex physical properties. For example, recent works have explored the perception of garments~\cite{chi2021garmentnets,xue2023garmenttracking}, fluids~\cite{lin2023pourit,narasimhan2022self}, and articulated objects~\cite{liu2022toward,li2020category}. Among these, articulated objects pose a unique challenge due to their multiple rigid kinematic parts, making their perception and manipulation particularly complex.
Inaccurate perception of articulated objects can lead to the robot damaging delicate joints, unlike with liquids and garments, which are less susceptible to damage due to their flexible nature.  In this work, we focus on advancing the perception and estimation of articulated objects to enhance robotic manipulation capabilities.

\begin{figure*}[htbp]
    \centering
    \includegraphics[width=0.9\textwidth]{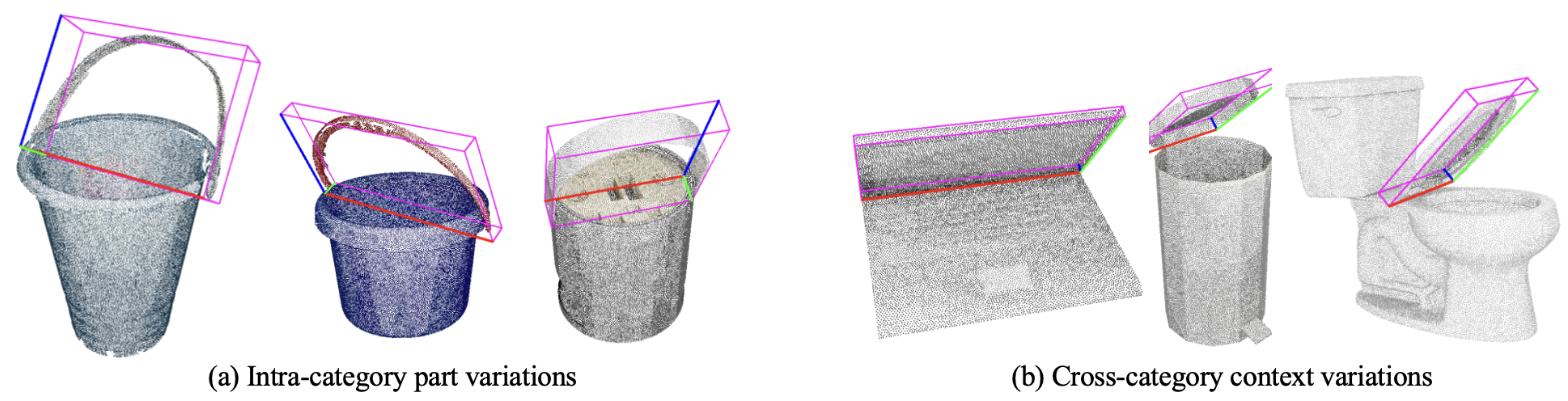}
       \vspace{-0.1in}
    \caption{(a) Illustration of challenge posed by intra-category part variations. Estimating the handles of different bucket types requires finding shared representations that can generalize across various instances within a category. 
(b) Illustration of challenge posed by cross-category context variations, exemplified by the variability of a hinge lid, which can be part of a laptop or a bin.This variability complicates the estimation of the pose and size of target parts across different categories of objects.  \label{fig:variation2} }
   \vspace{-0.2in}
\end{figure*}

However, there are still significant challenges in perceiving articulated parts. These challenges include: 
1) \textit{Intra-category part variations.} Novel articulated objects often lack exact 3D CAD models, necessitating intra-category generalization. For instance, estimating the handles of different bucket types requires finding shared representations that can generalize across various instances within a category, as illustrated in Fig.~\ref{fig:variation2} (a).
2) \textit{Cross-category context variations.} Articulated parts of a category exhibit vast variations in part contexts across different object categories. Unlike category-level rigid object pose methods~\cite{lin2022sar,zhang2024generative,chen2021fs}, which deal with single, consistent shapes, articulated objects have multiple kinematic parts leading to diverse contexts. Thus, even parts from the same instance can be assembled differently with other rigid parts. For example, a hinged lid can be part of a laptop or a bin, as shown in Fig.~\ref{fig:variation2} (b). This variability complicates the estimation of the pose and size of target parts across different categories of objects.
These challenges highlight the need for advanced methods to accurately perceive and manipulate articulated objects, accommodating both intra-category part variations and cross-category context differences.

To tackle these challenges, previous methods~\cite{li2020category,geng2023gapartnet} propose Normalized Part Coordinate Space (NPCS) representation to provide a normalized space for canonical part references. This representation maps instances within the same category into a canonical space, facilitating the learning of category-level mapping from the camera frame to a shared frame. For example, Li et al.~\cite{li2020category} address intra-category articulated pose estimation, but their method does not generalize to cross-category objects. Additionally, GAPartNet~\cite{geng2023gapartnet} learns domain-invariant features to facilitate the cross-category generalization. However, their method is a two-stage process, first segmenting the parts and then estimating the NPCS for each part individually. This cascaded pipeline accumulates segmentation errors, thereby reducing the accuracy of the NPCS estimation.

To address these challenges, this paper presents a simple yet efficient single-shot pipeline that You Only Estimate Once (YOEO), which provides a \textit{unified, one-stage,
real-time category-level} articulated object 6D Pose
estimation for robotic grasping. Particularly,
(1) To tackle the challenges of intra-category part variations, we represent each category of part in NPCS similar to works~\cite{li2020category,geng2023gapartnet}. This standardized space normalizes the position and orientation of parts, establishing a consistent reference frame for objects within the same category.  Consequently, this facilitates accurate 6D pose and size estimation for unseen parts.
(2) To address cross-category context variations, we jointly model semantic understanding and instance centroid offset while learning NPCS mapping. Semantic supervision enables the model to learn to distinguish between part classes belonging to the same object, thereby developing a more unified feature representation for each part class. Then, centroid offset learning enables the model to distinguish between multiple part instances within the same part classes. This enables the method to localize segments with similar category-level features in novel objects, even when these objects have different contextual parts, thus can generalizable to novel objects.

Typically, given that the point cloud of an articulated object contains multiple kinematic parts, we employ the unified network RandLA-Net~\cite{hu2020randla} to learn object features, facilitating simultaneous optimization of rigid part semantic segmentation, dense coordinate predictions in each NPCS map, and instance centroid offsets. This end-to-end optimization process enhances the performance of each output, thereby improving the accuracy of pose estimation. Subsequently, part semantic segmentation and instance centroid offsets are used to filter and cluster for instance segmentation. Once each instance is obtained, we further extract the region of estimated NPCS and register it with the point cloud using the Umeyama algorithm~\cite{umeyama1991least}, allowing the calculation of the final part pose and size.

In summary, the main contributions of this paper are as follows:
(1) We introduce a synthetic-to-real pipeline designed to perceive previously unseen articulated object instances from a single depth input in real-world settings.
(2) We propose an end-to-end unified network that concurrently estimates semantic labels, instance centroid offsets, and NPCS representations. This holistic optimization approach improves the accuracy of NPCS estimation and facilitates the generation of accurate 6D poses for each part, even amidst noisy depth data.
(3) Our method is deployed within a real-time robotic system, enabling the visual perception of articulated objects at a rate of 200Hz. Furthermore, it guides the robot in real-time manipulation of the target part, demonstrating practical applicability in dynamic environments.

\section{Related Work}
\textbf{3D Part-wise Objects Assets.}
The task of 3D part-wise object representation and manipulation has gained significant attention in robotics and computer vision. Large-scale 3D datasets are the cornerstone of research in this field, such as ShapeNet~\cite{chang2015shapenet}, Objaverse~\cite{deitke2023objaverse, deitke2024objaverse}, OmniObject3D~\cite{wu2023omniobject3d}, etc. However, merely having a holistic perception of objects is insufficient. For fine-grained robotic manipulation (e.g., opening bottle caps, pressing buttons and opening refrigerators), a focused perception of object parts is often required, which necessitates the support of 3D part-wise datasets. Many previous works~\cite{mo2019partnet, guo2020deep, downs2022google,li20223d} have abstracted the shapes of 3D objects and decoupled the parts to construct datasets, thereby promoting a series of studies on part-wise object perception~\cite{paschalidou2021neural, xu2022partafford, yang2021unsupervised}. Furthermore, a dataset capable of supporting cross-category domain-generalizable object perception, GAPartNet~\cite{geng2023gapartnet}, with rich part annotations, offering valuable guidance. Specifically, our model was primarily trained and tested on ~\cite{geng2023gapartnet}, and demonstrated its ability to estimate the part poses of the articulated objects.

\begin{figure*}[htbp]
    \centering
    \includegraphics[width=1.0\textwidth]{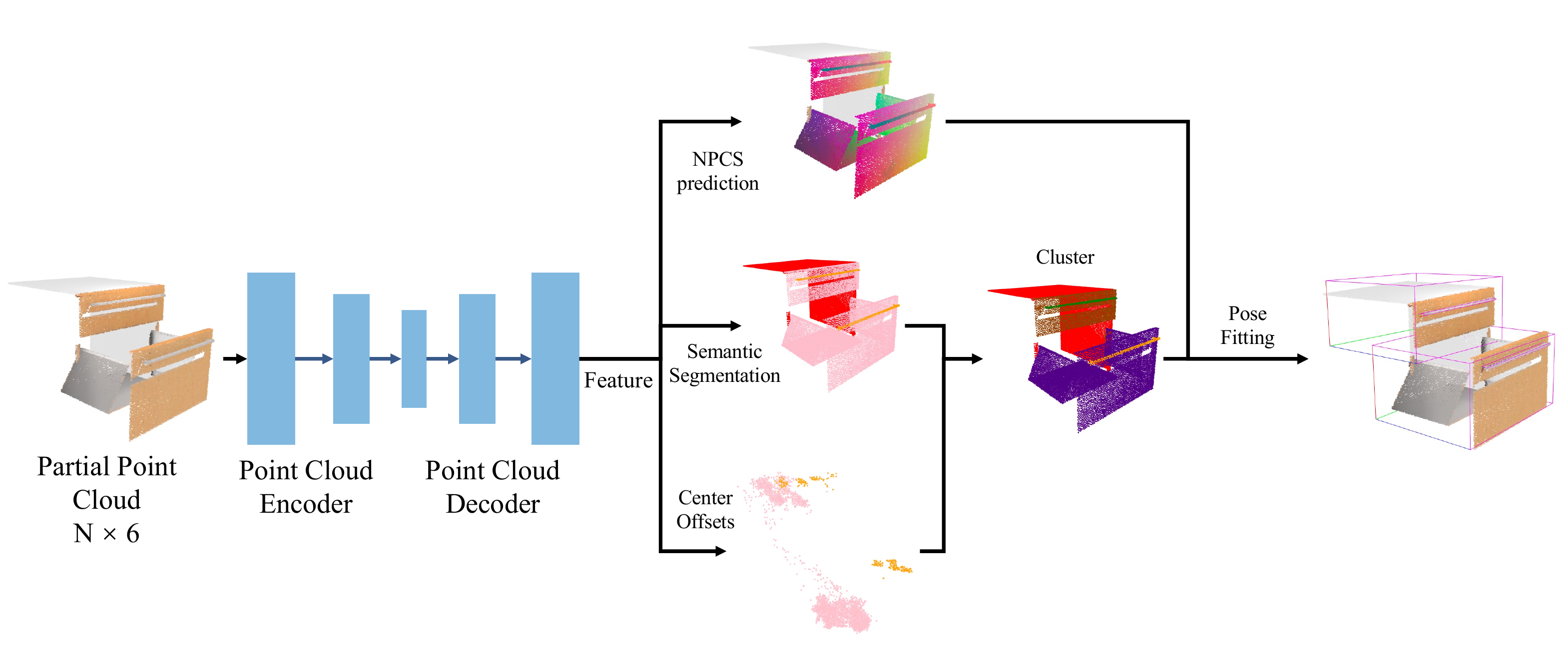}
    \vspace{-0.15in}
    \caption{\textbf{Architecture overview.} The Feature Extraction module extracts the per-point feature from an partial point cloud. They are fed into three
parallel modules to predict the NPCS maps, semantic labels and the offsets to centroids  of each point. A
clustering algorithm is then applied to distinguish different instances with the same semantic label and points on the same instance. Finally, an aligning algorithm is applied to the predicted npcs map and real point cloud to estimate 6DoF pose parameters.   \label{fig:pipeline} }
\vspace{-0.2in}
\end{figure*}

\textbf{Part Instance Segmentation and Clustering from Point Cloud Observations.}
Part instance segmentation in 3D point cloud is a challenging task due to the irregular and sparse nature of the data. Existing methods such as PointNet++~\cite{qi2017pointnet++} and SGPN~\cite{wang2018sgpn} utilize deep learning techniques to segment instances by learning point-wise features. However, these methods typically require a separate clustering step to group points into instances, which can be computationally expensive. Recent advancements like VoteNet~\cite{ding2019votenet} and 3D-SIS~\cite{hou20193d} have improved clustering efficiency but still involve multi-stage processes. Our approach integrates instance segmentation and clustering~\cite{he2020pvn3d, he2021ffb6d} within a unified network, leveraging point-wise centroid offsets to facilitate efficient and accurate segmentation. This end-to-end learning framework not only simplifies the pipeline but also improves the segmentation quality and speed.

\textbf{Category-level  Rigid Object Pose Estimation.}
Rigid object pose estimation deals with objects that maintain a fixed shape and structure, necessitating the determination of a single, static pose in 3D space. Some works~\cite{cheang2022learning, sun2023language, wang2024polaris, wang2023wall, wang2019normalized, zhang2024generative, lin2022sar} estimate the pose and size from single view RGB-D images. For example, NOCS~\cite{wang2019normalized}
Some point-based methods like FS-Net~\cite{chen2021fs}, SAR-Net~\cite{lin2022sar}  and GenPose~\cite{zhang2024generative} focus on estimated the pose by learning the geometry shape of the instances. However, these methods are only suitable for rigid bodies, limiting their potential to be extended for perceiving complex objects composed of multiple movable parts.

\textbf{Category-level Articulated Object Pose Estimation.} Conversely, articulated object pose estimation addresses objects composed of multiple interconnected parts that can move relative to each other, requiring the estimation of both the overall pose and the configuration of individual movable components.To enhance accuracy and generalization across unseen 
articulated objects, Articulation-aware Normalized
Coordinate Space Hierarchy (ANCSH)~\cite{li2020category} was proposed to represent different articulated objects in a given category.~\cite{gadre2021act} uses interactive learning to segment articulated objects into parts, discovering structures effectively and generalizing to unseen categories. 
AKB-48~\cite{liu2022akb} project offers a comprehensive Articulated object Knowledge Base with 2,037 real-world 3D models, supported by a fast modeling pipeline. GAPartNet~\cite{geng2023gapartnet} introduces a two-stage method for domain-generalizable 3D part segmentation and pose estimation by learning domain-invariant features. However, this two-stage pipeline has slow inference speed and tends to accumulate errors from the segmentation stage.
\section{Method}
\noindent \textbf{Task Formulation.} Given the point cloud $\mathcal{P} \in \mathbb{R}^{N\times3}$ of the articulated object, our task is to estimate the semantic labels $C_i$, the normalized object part coordinate maps $M_i$, and the centroid offsets $O_i$ for the $i$-th point. We first utilize the semantic labels to cluster different part classes. Subsequently, we cluster based on the centroid offsets $O_i$ to differentiate between different part instances that share the same semantic label. Utilizing the normalized object part coordinate maps $M_i$, we determine the NPCS map for each part instance. Once the point cloud $\mathcal{P}$ is available, we register the estimated NPCS map with the corresponding points to calculate the transformation parameters $\{s, R, t\} \in SIM(3)$, where $s \in \mathbb{R}$, $R \in SO(3)$, and $t \in \mathbb{R}^3$. $SIM(3)$ is the Lie group of 3D similarity transformations.

\noindent \textbf{Architecture Overview.}
{As shown in Fig.\ref{fig:pipeline}, our network processes the input point cloud, which is obtained from the output of the segmentation model. Here, we use Grounding-DINO\cite{liu_grounding_2023},which is a vision-language model that detects and segments objects based on textual descriptions, to generate the input point cloud. The network then estimates semantic class labels, NPCS maps, and centroid offsets for each point simultaneously.} Clustering based on these centroids groups points belonging to the same instance. Part labels are then assigned to each instance to locate the filtered NPCS maps of each part. Finally, the transformations and scales between the actual points and the estimated NPCS maps yield the 6-DoF pose and 3D size of each part.

\begin{figure}
    \centering    \includegraphics[width=0.5\textwidth]{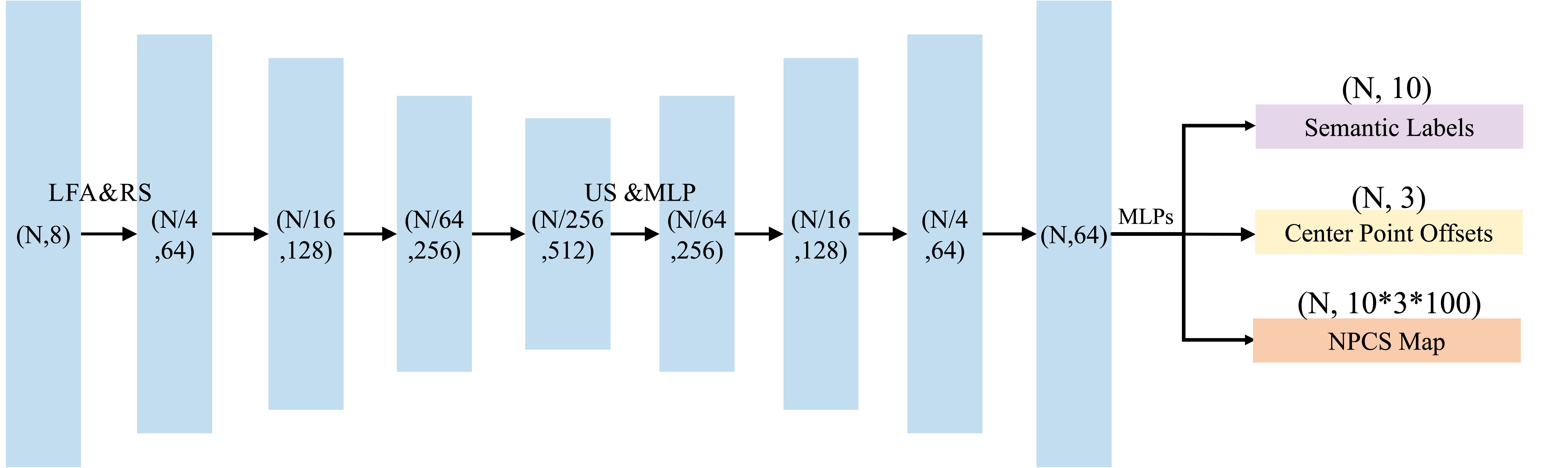}
    \caption{\textbf{The detailed architecture of our YOEO.} FC: Fully Connected layer, LFA: Local
Feature Aggregation, RS: Random Sampling, MLP: shared Multi-Layer Perceptron, US: Up-sampling.   \label{fig:detail_pipe} }
    \vspace{-0.2in}   
\end{figure}

\noindent \textbf{Details of the Network.} Specifically, our point-based method consists of an encoder and decoder module. The details of each module are shown in Figure~\ref{fig:detail_pipe}. We use RandLA-Net~\cite{hu2020randla} for feature extraction from the point cloud. The extracted features from the network are fed into the following modules: semantic segmentation, center point offset prediction, and NPCS Map prediction, all of which are composed of shared MLPs.

\subsection{Semantic Part Learning}

To handle object categories with multiple parts, previous methods utilize existing grouping architectures to process features extracted from the backbone network, using a post-processing module to obtain part segmentation masks. Tian et al.~\cite{tian_shape_2020} build the
pose estimation models with the segmentation masks as input to simplify the problem. The pose estimation problem
 is divided into two stages, part segmentation and part pose estimation, which are trained separately. However, by incorporating the part pose estimation problem into the Semantic Part Learning module, the NPCS learning module, and the Center Offset Learning module, we hypothesize that these three tasks can enhance each other’s performance through parallel training. Our ablation study confirms this hypothesis. Firstly, the semantic segmentation
module forces the model to extract global and local features
on each instance to distinguish different part classes, which allows the NPCS learning module to focus more on the part. Secondly, the semantic segmentation provides distinct semantic labels, which helps the Centroid Offset Learning module more accurate in distinguishing the centroids of different parts, as different semantic labels correspond to different centroids.

Based on this observation, we introduce a pointwise
part semantic segmentation module $\mathcal{M}_s$ into the
network and jointly optimize it with module $\mathcal{M}_c$ and $\mathcal{M}_n$ . 
Specifically, the semantic segmentation module $\mathcal{M}_s$ predicts semantic labels for each point by using the extracted features. The supervision for this module is provided using Focal Loss\cite{lin2017focal}.
\begin{equation}
L_{\text{semantic}} = -\alpha (1 - q_i)^\gamma \log(q_i) \quad \text{where} \quad q_i = c_i \cdot l_i
\end{equation}
Here, $\alpha$ and $\gamma$ are the balancing and focusing parameter, respectively; $c_i$ is the predicted confidence for the $i$-th point belonging to a specific class, and $l_i$ is the one-hot encoded ground truth class label.

\subsection{Centroid Offset Learning}

Considering that there can be multiple part instances with the same semantic label in an object, we design the Centroid Offset Learning module to predict the centroid of each instance to distinguish between them.
It utilizes the per-point feature to predict the Euclidean translation offset $\Delta x_i$ to the associated object center. The learning process of $\Delta x_i$ is guided by an L1 loss:
\begin{equation}
\label{eqn:Lctr}
    L_{\textrm{center}} = \frac{1}{N} \sum_{i=1}^{N} ||\Delta x_i - \Delta x_i^* || \mathbb{I}(p_i \in I)
\end{equation}
In this equation, $N$ represents the total number of seed points on the object's surface, and $\Delta x_i^*$ is the ground truth translation offset from seed $p_i$ to the instance center. The indicator function $\mathbb{I}$ specifies whether point $p_i$ belongs to the particular instance.

\subsection{NPCS learning for pose and size estimation}

For the Normalized Part Coordinate Space Map Learning module, we aim to learn a mapping $\Phi: \mathcal{P}_{o} \rightarrow \mathcal{P}_{\mathbb{C}}$, where $\mathcal{P}_{o}$ represents the observed object point cloud and $\mathcal{P}_{\mathbb{C}}$ represents the canonical space point cloud. 
Both $\mathcal{P}{o}$ and $\mathcal{P}_{\mathbb{C}}$ consist of 3 channels, representing the 3D coordinates. $\Phi(\cdot)$ is constructed using a PointNet-like architecture for its lightweight design and computational efficiency \cite{qi2017pointnet}. The learning task is formulated as a classification problem by discretizing the coordinates $p^{i}_{\mathbb{C}}$ into 100 bins for each of the three axes (x, y, and z).
 For each region filtered by the predicted part segmentation mask \( C_i \), we use the Softmax cross-entropy loss, as it has proven to be more effective than regression in reducing the solution space
\cite{wang2019normalized}. In addition to the predicted dense correspondence, the 6D object pose $\xi_{o}\in \{SE(3)\}$ is also recovered. This is computed using \textit{RANSAC} for outlier elimination and the \textit{Umeyama algorithm}\cite{fischler1981random} to determine the transformation parameters $\{s, R, t\} \in SIM(3)$ from the predicted canonical space point cloud $\mathcal{P}_{\mathbb{C}}$ to the observed object segment point cloud $\mathcal{P}_{o}$, ensuring that the rotation component is orthonormal.
{\subsection{Grasping, Manipulation Strategy and Motion policy}}

Utilizing the NPCS representation, we possess information about the joint or prismatic axis in the NPCS frame, along with predefined category-level grasp poses. By aligning the NPCS with the real-world point cloud through registration, we can transform both the actionable axis and predefined grasp poses from the NPCS frame to the camera frame. In real robot experiments, the camera is calibrated to the robot’s base frame, enabling a straightforward transformation of these elements into the robot frame for motion planning.

We also define category-level motion policies within the NPCS framework. During actual manipulation, aligning the NPCS with the real-world point cloud allows us to transform motion policies from the NPCS frame to metric space. This approach ensures that our system not only adheres to the theoretical framework but also adapts effectively to real-world physical constraints, such as variations in object size and position.

\begin{table*}[htbp] \small
\centering
\footnotesize
\caption{\textbf{Results of Part Pose Estimation in terms of \boldmath{$R_{e}$} ($^{\circ}$), \boldmath{$T_{e}$} (cm), \boldmath{$S_{e}$} (cm), \textbf{mIoU}=\textbf{3D mIoU} (\%), $\textbf{A}_5$=\textbf{5$^{\circ}$5cm accuracy} ($\%$), $\textbf{A}_{10}$=\textbf{10$^{\circ}$10cm accuracy} ($\%$), \textbf{Param} (millions) and \textbf{Speed} (Hz).} PG=baseline modified from PointGroup \cite{jiang2020pointgroup}. AGP=baseline modified from AutoGPart\cite{liu2022autogpart}.\label{tab:pose_evel}}
\begin{tabular}{c|cccccccc}
\toprule
Method & {\boldmath{$R_{e}$}} $\downarrow$ & {\boldmath{$T_{e}$}}$\downarrow$ & {\boldmath{$S_{e}$}}$\downarrow$ & \textbf{mIoU} $\uparrow$ & $\textbf{A}_{5}$ $\uparrow$ & $\textbf{A}_{10}$ $\uparrow$ & \textbf{Param.(M)}$\downarrow$ & \textbf{Speed (Hz)}$\uparrow$\\
\midrule[0.5pt]
PG\cite{jiang2020pointgroup}  & 14.3 & 0.034 & 0.039 & 49.4 & 24.4 & 47.0 & / & / \\ 
AGP\cite{liu2022autogpart}  & 14.4 & 0.036 & 0.039 & 48.7 & 26.8 & 49.1 & / & /\\
GAPartNet\cite{geng2023gapartnet}  & 9.9 & \textbf{0.024} & \textbf{0.035} & 51.2 & 28.3 & 53.1 & 7.9 &20\\
Ours  & \textbf{9.0} & 0.11 & 0.036 & \textbf{57.6} & \textbf{30.4} & \textbf{54.4} & \textbf{1.9} & \textbf{200}\\ 
\midrule[0.5pt]
\end{tabular}
\end{table*}

\section{Experiment}
\noindent\textbf{GAPartNet Dataset.} 
The GAPartNet dataset is a comprehensive resource designed to facilitate research in articulated object manipulation. It encompasses 9 distinct classes of parts, each accompanied by detailed semantic labels and pose annotations. The dataset includes a total of 8,489 part instances derived from 1,166 objects, which span 27 diverse object categories. On average, each object within the dataset has 7.3 functional parts, highlighting the complexity and variety of the dataset.
A notable characteristic of GAPartNet is its extensive cross-category representation: each class of parts appears in objects from at least 3 different object categories, and on average, a single part class is represented across 8.8 object categories. This diverse cross-category distribution is pivotal for establishing a robust benchmark for evaluating and enhancing generalizable part recognition and pose estimation methods.

\noindent\textbf{Evaluation Metric.}
We evaluate part pose estimation performance using metrics such as average rotation error \( R_e (^\circ) \), translation error \( T_e (\text{cm}) \), scale error \( S_e (\text{cm}) \), and translation error of the part interaction axis \( d_e (\text{cm}) \). Specifically, we follow the standards of the GAPartNet Dataset, where the scales of all objects are normalized to a range of 0 to 1 cm. Additionally, we measure 3D Intersection over Union (3D mIoU) and accuracy percentages for specific thresholds: \( 5^\circ, 5\text{cm} \) and \( 10^\circ, 10\text{cm} \). Furthermore, the parameters of the networks and inference speeds, which are calculated from feeding object point clouds to get part poses, are considered. 

\subsection{Comparison to Baselines}
\begin{figure}
    \centering
    \includegraphics[width=0.5\textwidth]{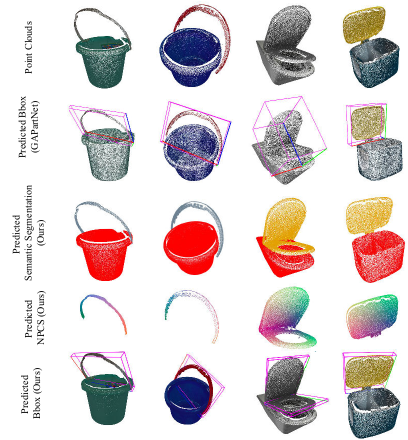}
        \vspace{-0.1in}   
    \caption{Qualitative results on the GAPartNet dataset. The left two columns illustrate the intra-category results for hinge handles within the bucket category. The right two columns display the cross-category results for hinge lids across toilet and box categories.    \label{fig:pic_v3} }
 
    \vspace{-0.15in}   
\end{figure}

\begin{figure}
    \centering
    \includegraphics[width=0.5\textwidth]{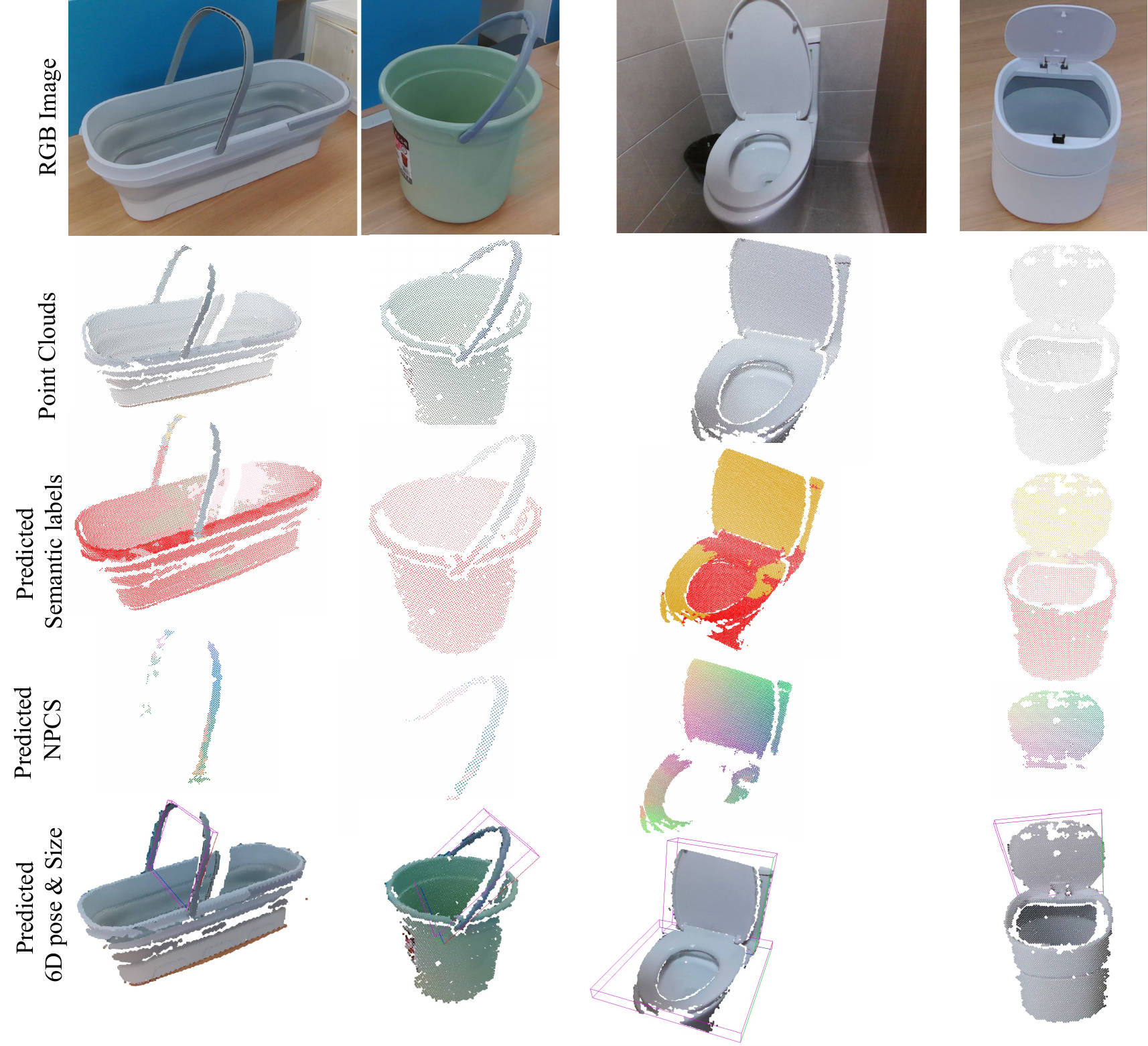}
    \caption{Qualitative results of the real-world perception by using our YOEO method. We captured the object's RGB images and point cloud using~\cite{MantisVision}. The left two columns illustrate the intra-category results for hinge handles within the bucket. The right two columns show the cross-category results for hinge lids across toilet and box classes.  \label{fig:real_world} }
\end{figure}

We compared the pose estimation accuracy, inference speed, and model parameters with the baseline methods, and the summarized results are presented in Table~\ref{tab:pose_evel}. Our method demonstrates significantly improved pose accuracy compared to the previous state-of-the-art method, GAParNet, particularly in the mIoU metric, validating the accuracy of the estimated poses. In comparison to the two-stage method GAParNet, our approach requires fewer parameters and achieves faster inference speeds, thereby reducing computational cost and enabling deployment on devices with limited computational resources. We also visualize the qualitative results in Fig.~\ref{fig:pic_v3} and Fig.~\ref{fig:real_world}.

{\subsection{Ablation study}}
We trained each of the three prediction heads individually by freezing the other two, repeating the process three times, once for each head. Then, we combined the individually trained heads and compared the results to those obtained from co-training. The results clearly support our conjecture that combining the prediction heads enhances the overall performance of our model compared to training them separately.

As shown in Table~\ref{tab:abalation}, parallel (co-) training consistently improves performance across all metrics. The rotation error ($R_e$) drops from 19.6 to 9.0, while the translation error ($T_e$) and scale error ($S_e$) are reduced from 0.14 to 0.11 and from 0.041 to 0.036, respectively. 3D Intersection over Union (3D mIoU) also improves significantly, increasing from 52.3\% in individual training to 57.6\% in parallel training. These results indicate that parallel training leads to more accurate pose and size estimation.

\begin{table}[h] \small
\centering
\footnotesize
\caption{Ablation Study: Individual vs. Parallel Training \label{tab:abalation}}
\begin{tabular}{c|cccccccc}
\toprule
Method & {\boldmath{$R_{e}$}} $\downarrow$ & {\boldmath{$T_{e}$}}$\downarrow$ & {\boldmath{$S_{e}$}}$\downarrow$ & \textbf{mIoU} $\uparrow$ & $\textbf{A}_{5}$ $\uparrow$ & $\textbf{A}_{10}$ $\uparrow$\\
\midrule[0.5pt]
Ind. Training & 19.6 & 0.14 & 0.041 & 52.3 & 23.9 & 52.4 \\
Para. training  & \textbf{9.0} & \textbf{0.11} & \textbf{0.036} & \textbf{57.6} & \textbf{30.4} & \textbf{54.4} \\ 
\midrule[0.5pt]
\end{tabular}
\end{table}

\subsection{Robotic Experiment}
\noindent\textbf{Hardware Settings.} Our algorithm is deployed on a PC workstation equipped with an Intel i9-13900K CPU and an NVIDIA RTX 6000 Ada Generation GPU to provide visual perception of the target objects. To execute the grasping and manipulation tasks, we utilize the Kinova Gen2 6-DoF robotic arm. This robotic arm features three under-actuated fingers, each of which can be individually controlled. A MantisVision camera \cite{MantisVision} is used to capture RGB-D images of the scene and is mounted on a tripod positioned opposite the robot workspace. The camera is calibrated to the robotic base frame.

\noindent\textbf{Task Description.} 
To assess the sim-to-real capability of our method and evaluate its robustness and generalizability, we deployed our algorithm on a real robotic arm, specifically the KINOVA robot arm.To ensure the representativeness of our experiments, we selected three distinct part classes: drawer, hinge lid, and hinge handle. The corresponding tasks involved pulling the drawer, lifting the lid, and raising the handle.

\noindent\textbf{Evaluation Metric.} Depending on the specific experimental task, different metrics were used. For the drawer task, the robot arm successfully completed the task by pulling the drawer out 0.2 meters. For the hinge handle task, success was defined by rotating the handle 30 degrees around its axis. Similarly, for the hinge lid task, the robot arm successfully completed the task rotating the lid 50 degrees around its axis.

\noindent\textbf{Results.}
The success rate of manipulating articulated objects in real-world robotic experiments is summarized in Table~\ref{tab:comparison}. The results show that our lightweight model competes effectively with the baseline method, GAParNet. Our single-shot approach accurately generates poses that guide the robot in interacting with objects not seen during the training stages, demonstrating the utility of our method in robotic applications.

\begin{table}\small
    \centering
    \caption{Robot Manipulation Success Rate. \label{tab:comparison}}
    \begin{tabular}{ccccccc}
        \toprule
        & hinge handle & drawer & hinge lid & Total \\
        \hline
        GAPartNet \cite{geng2023gapartnet} & 7/10 & \textbf{7/10} & 6/10 & 20/30\\
        \midrule[0.5pt]
        Ours & \textbf{9/10} & 5/10 & \textbf{8/10} & \textbf{22/30} \\
        \bottomrule
    \end{tabular}
    \vspace{-0.1in}
   
\end{table}


\section{Conclusion}

We present YOEO, a lightweight model for real-time category-level articulated object 6D pose estimation. Unlike multi-stage methods, YOEO employs a single-stage framework directly on the point cloud, enabling end-to-end part pose estimation. It efficiently combines instance segmentation and NPCS representations, utilizing accurate point offset calculations and clustering for precise NPCS region alignment. Experiments on the GAPart dataset and real-world data demonstrate its real-time synthetic-to-real pose estimation capability. Robotic experiments on Kinova Gen 2 further showcase its proficiency with unseen articulated objects.


\textbf{Limitations}. 
Our method for articulated object pose estimation faces two challenges: suboptimal performance on smaller objects and inaccuracies with metallic surfaces due to poor point cloud quality. Future work will integrate RGB information for improved precision.

\section{ACKNOWLEDGMENT}
This work is supported in part by NSFC Project (62176061), Shanghai Municipal Science and Technology Major Project
(No.2021SHZDZX0103), and Shanghai Technology Development and Entrepreneurship Platform for Neuromorphic and AI SoC.









\bibliographystyle{IEEEtran}
\bibliography{example}

\end{document}